\documentclass{article}
\usepackage{spconf,amsmath,epsfig}

\usepackage{bbm}
\usepackage{subfigure}
\usepackage{algorithm}
\usepackage{algorithmic}
\usepackage{amssymb}
\usepackage{multicol}
\usepackage{multirow}
\usepackage[pagebackref=true,breaklinks=true,letterpaper=true,colorlinks,bookmarks=false,citecolor=blue]{hyperref}
\usepackage[numbers,sort&compress]{natbib}
\usepackage{lipsum}

\title{Accurate Building Detection in VHR Remote Sensing Images \\ using Geometric Saliency}
\name{Jin Huang$^1$, Gui-Song Xia$^1$, Fan Hu$^1$, Liangpei Zhang$^1$}
\address{$^1$State Key Lab. LIESMARS, Wuhan University, Wuhan, China}
\begin{document}
	%\ninept
	%
	\maketitle
	
	\newcommand\blfootnote[1]{% 
		\begingroup 
		\renewcommand\thefootnote{}\footnote{#1}% 
		\addtocounter{footnote}{-1}% 
		\endgroup 
	}
	%
	%\vspace{-5mm}
	\begin{abstract}
		This paper aims to address the problem of detecting buildings from remote sensing images with very high resolution (VHR).
		Inspired by the observation that buildings are always more distinguishable in geometries than in texture or spectral, we propose a new geometric building index (GBI) for accurate building detection, which relies on the geometric saliency of building structures.
		The geometric saliency of buildings is derived from a mid-level geometric representations based on meaningful junctions that can locally describe anisotropic geometrical structures of images.
		The resulting GBI is measured by integrating the derived geometric saliency of buildings.
		Experiments on three public datasets demonstrate that the proposed GBI achieves very promising performance, and meanwhile shows impressive generalization capability.\blfootnote{This work is supported by NSFC projects under the contracts No.61771350 and No.41501462.}
		
		%This paper aims to address the problem of detecting buildings from remote sensing images with very high resolutions (VHR). Inspired by the observation that buildings are always more distinguishable in geometries than in texture or spectral, we present a new geometric building index (GBI) for accurate building detection, which relies on the geometric saliency of building structures. We derive the geometric saliency of buildings from a mid-level geometric representations including meaningful junctions that can locally describe anisotropic geometrical structures of images. The resulting GBI is measured by integrating the computed geometric saliency bounded in a parallelogram.
		%Experiments on three public datasets demonstrate that without any training samples, the proposed GBI achieves very promising performance, and meanwhile shows impressive generalization capability.
	\end{abstract}

	\vspace{-3mm}
	\section{Introduction}
	\label{sec:intro}
	\vspace{-2mm}
	Accurate building maps play an important role in a wide range of applications, such as urban planning and 3D city modeling.
	Nowadays, the large amounts of increasingly available remote sensing (RS) images with very high resolution (VHR) up to half a meter provide abundant data sources to generate such accurate building maps.
	However, manual administration of buildings from huge volume of VHR-RS images is unfeasible, hence there is an urgent demand to develop automatic approaches for detecting buildings from VHR-RS images.
	
	Over the past years, many studies have been devoted to automatic building detection,  e.g.,~\cite{zha_use_2003-1,pesaresi_robust_2008,shao_basi:_2014,sirmacek_urban_2009,sirmacek_urban_2010,huang_mbi_2012,saito_multiple_2016}.
	Among them, one main stream exploits the discriminative properties of buildings in RS images, e.g., from the aspects of spectrum~\cite{zha_use_2003-1}, texture~\cite{pesaresi_robust_2008,shao_basi:_2014} and local structural or morphological features~\cite{sirmacek_urban_2009,sirmacek_urban_2010,huang_mbi_2012}.
	These methods perform well on detecting buildings from mid-/high-resolution RS images, but dramatically lose their efficiency for RS images of half-meter resolution.
	The performance decrease is largely due to the fact that, in VHR-RS images, textural or spectral information lacks discriminative power to distinguish buildings.
	Moreover, most of these approaches are incapable of providing accurate boundaries of buildings, which are particularly desirable in the precise mapping of buildings.
	Another stream of the state-of-the-art building detection approaches attempts to detect buildings by learning an off-the-shelf parameterized model, e.g., the convolutional neural networks (CNNs), with manually labeled samples~\cite{saito_multiple_2016,zuo_hf-fcn:_2016}.
	Despite the high performance of learning-based methods, especially the ones based on CNNs~\cite{zuo_hf-fcn:_2016}, their performances heavily rely on a considerable amount of well-annotated training samples, and thus they have very limited generalization capability beyond the training domain.
	
	This paper presents a new method for accurately detecting buildings in VHR-RS images, by computing the geometric saliency of building structures.
	Our work is inspired by the observation that, in VHR-RS images, buildings are always more distinguishable in geometries (both local and global) than in texture or spectrum.
	More precisely, we first propose to represent VHR-RS images with a mid-level geometrical representation, by exploiting junctions that can locally depict anisotropic geometrical structures of images.
	We then derive the saliency of geometric structures on buildings, by considering both the probability of each junction that measures its saliency to its surroundings and the relationship of junctions.
	This stage can encode both local and semi-global geometric saliency of buildings in images.
	Finally, the geometric building index (GBI) of whole image is measured via integrating the derived geometric saliency.
	
	In contrast to existing building indexes, e.g.~\cite{zha_use_2003-1,pesaresi_robust_2008,shao_basi:_2014,sirmacek_urban_2009,sirmacek_urban_2010,huang_mbi_2012}, our method results in less redundant non-building areas and can provide accurate contours of buildings, thanks to the geometric saliency computed from a mid-level geometrical representation.
	As we shall see in Section \ref{sec:experiment}, our method achieves the state-of-the-art performance\footnote{All results are available at \url{http://captain.whu.edu.cn/project/geosay.html}.} on building detection and meanwhile shows promising generalization power to different datasets, especially in comparison with learning-based approaches~\cite{zuo_hf-fcn:_2016}.
	
	\vspace{-3mm}
	\section{Methodology}
	\label{sec:index}
	
	\begin{figure*}[htb!]
		\centering
		\includegraphics[width = 0.95\linewidth]{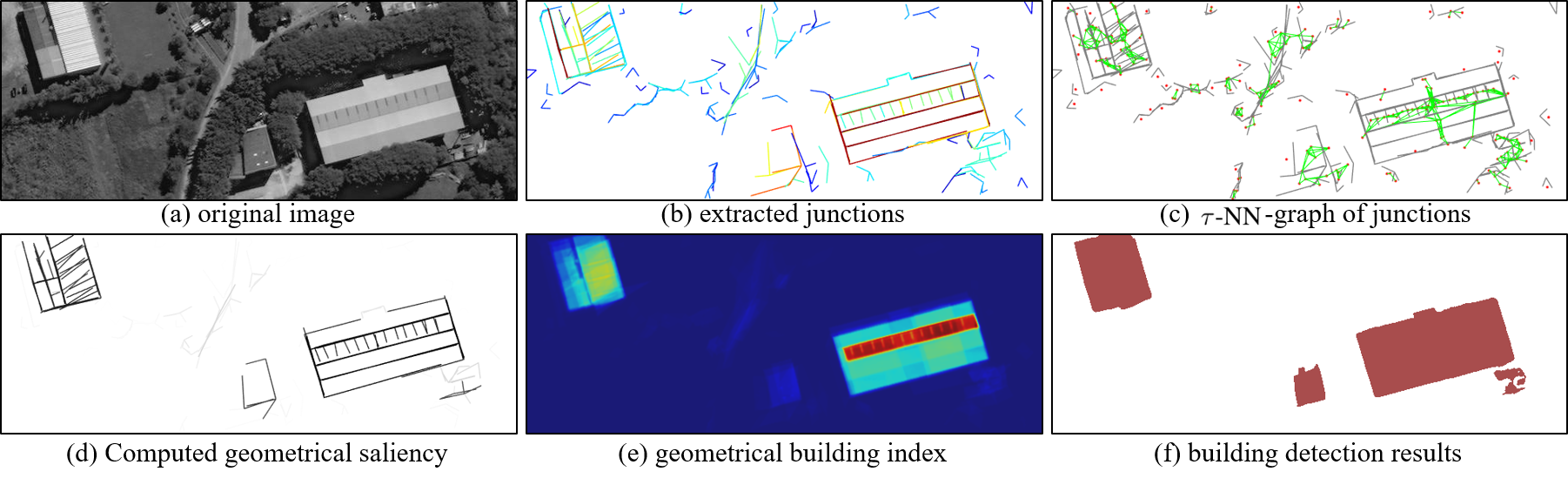}
		\vspace{-3mm}
		\caption{A running example on building detection with geometric saliency.
			%Left top: structure of L-junction. Left bottom: L-junctions located in the corners of a rectangular building could cover most of the building. Right: image with junctions detected. Colorful lines represent junctions and their reliability (increasing from blue to red).
		}
		\vspace{-3mm}
		\label{fig:junc_detail}
	\end{figure*}
	
	\vspace{-3mm}
	\subsection{A mid-level geometric representation of images}
	Let $u: \Omega \mapsto \mathcal{Z}_+^L$ denote an $L$-channel VHR-RS image defined on the image grid $\Omega$.
	For imagery in panchromatic format, all the geometrical information is contained in the single channel image $u$.
	While, for a multi-spectral image $u = \{u_1, u_2, . . . , u_L\}$, the main geometrical structures of the image can be computed from its $p$-energy image $U = (\sum_{i=1}^L u_i^p)^{p^{-1}}$ or from its first PCA component~\cite{xia-stucture-2010}.
	In this work, we concentrate on dealing with satellite images with (R,G,B)-channels, so the analysis of geometrical information is based on the luminance channel, with $p=1$.
	
	This work proposes to use a mid-level geometric representation of VHR-RS images\cite{Xue2017Anisotropic}.
	For an image $U$, let
	$
	\mathcal{J}
	$
	denote all detected junctions,
	where each junction $\jmath \in \mathcal{J}$ is encoded as $\jmath : \{\mathbf{p}, \{\theta_i, s_i\}_{i=1}^{N}, \rho \}$. $\mathbf{p} =(x, y) \in \Omega$ is the location of $\jmath$. $\{\theta_i\}_{i=1}^{N}$ and $\{s_i\}_{i=1}^{N}$ are the orientations and lengths of its $N$ branches respectively. $0 \leq \rho \le 1$ is the significance measured by {\it number false alarms} associated with junction $\jmath$.
	These junctions can be well detected by the anisotropic-scale junction (ASJ) detector.
	An example of detected junctions is displayed in Fig. \ref{fig:junc_detail}~(b).
	
	Observe that the mid-level geometric description $\mathcal{J}$ includes junctions with different number of branches, e.g., $L$-, $Y$-, and $X$-junctions with $2, 3$ and $4$ branches respectively. According to empirical studies, in terms of buildings, junctions with more than $3$ branches are rare and we decompose all junctions into $L$-junctions, to develop a building-centric geometric representation. Thus, we rewrite the junction format as
	$$
	\jmath : \{\mathbf{c}, \vec \nu_1, \vec \nu_2, \rho \},
	$$
	where $\vec \nu_1, \vec \nu_2$ are the two branches of $L$-junctions and $\vec \nu_i = \overrightarrow{\mathbf{p}\mathbf{q}_i}$ with $\mathbf{q}_i = \mathbf{p} + s_i \cdot (\cos \theta_i, \sin \theta_i)^\top$ for $i=1,2$. $\mathbf{c}$ is the center of the L-junction $\jmath$, and $\mathbf{c} = \frac{(\mathbf{q}_1 + \mathbf{q}_2)}{2}$. The significance $\rho$ inherits from its original junctions.
	Fig. \ref{fig:junc_detail}~(c) displays all the L-junctions, illustrating their centers with red dots.

	\vspace{-3mm}
	\subsection{Computing geometric saliency in VHR-RS images}
	In order to detect buildings, we need to derive certain geometric saliency from the mid-level geometric representation of VHR-RS images, so as to highlight geometric features inside buildings and suppress those outside buildings.
	To this end, we use the significance from both single geometrical primitives and pair-wise junctions.
	
	\vspace{1mm}
	\emph{\bf First-order geometric saliency $\omega^{(1)}$} :
	For an image $u$, the significance $\rho$ of each junction $\jmath$ detected by the ASJ detector indicates the reliability of the junction $\jmath$ appearing in $u$.
	The smaller the $\rho$ is, the more salient detected junction will be.
	%$$
	%\omega^{(1)} = \rho_\jmath.
	%$$
	In addition, it is noted that all detected junctions $\mathcal{J}$ can be divided into two subsets, {\it i.e.,} $\mathcal{J} = \mathcal{J}_{B} \cup \mathcal{J}_{\bar B}$, $\mathcal{J}_B$ inside buildings and $\mathcal{J}_{\bar B}$ outside buildings.
	Given a junction $\jmath$ with parameters $\Theta_\jmath \doteq \{\mathbf{c}, \vec \nu_1, \vec \nu_2, \rho \} $, the posterior probability $\mathbb{P}(\jmath \in \mathcal{J}_B \, | \, \Theta_\jmath)$, measuring the possibility of the event that a junction $j$ parameterized by $\Theta_\jmath$ is inside buildings, is derived by
	$$
	\mathbb{P}(\jmath \in \mathcal{J}_B \, | \, \Theta_\jmath) = \frac{\mathbb{P}( \Theta_\jmath \,| \, \mathcal{J}_B ) \mathbb{P}(\mathcal{J}_B)}{\mathbb{P}( \Theta_\jmath \,| \, \mathcal{J}_B ) \mathbb{P}(\mathcal{J}_B) + \mathbb{P}( \Theta_\jmath \,| \, \mathcal{J}_{\bar B} ) \mathbb{P}(\mathcal{J}_{\bar B})},
	\label{eq:jbs-jtheta}
	$$
	where the prior probabilities $P(\mathcal{J}_B), \, P(\mathcal{J}_{\bar B})$ and the likelihoods $\mathbb{P}( \Theta_\jmath \,| \, \mathcal{J}_B )$ $\mathbb{P}( \Theta_\jmath \,| \, \mathcal{J}_{\bar B})$ can be estimated from a given dataset of buildings, e.g., the Spacenet65 dataset as we shall see in Section \ref{sec:experiment}.
	Thus the first-order geometric saliency of a junction $\jmath$ can be computed as
	\begin{align}
	\omega_\jmath^{(1)} = (1- \rho_\jmath) \cdot \mathbb{P}(\jmath \in \mathcal{J}_B \, | \, \Theta_\jmath).
	\end{align}

	\emph{\bf Pairwise geometric saliency $\omega^{(2)}$} :
	%Buildings always have several corners and many junctions would be detected on one building.
	When there are many junctions whose centers are very close to each other in a region, the probability of existence of a building (building saliency) will be higher. Thus, pair-wise relationships of junctions are useful cues to derive geometric saliency. In contrast with first-order saliency, pair-wise ones can encode more globally geometric information in images.
	Here, we use nearest neighbors to compute pair-wise saliency.
	For a junction $\jmath$, its $\tau$-{\it nearest neighbors} ($\tau$-NN), denoted by $\mathcal{N}_\jmath$, are defined as a set of junctions satisfying
	$$
	\| \mathbf{c}_\jmath - \mathbf{c}_{\jmath'} \|_2 < \tau, \, \forall \jmath' \in \mathcal{N}_\jmath,
	\label{eq:distance-constraint}
	$$
	where $\tau$ represents the minimal length of branches of the junction $\jmath$. An example of the $\tau$-NN graph for junctions is displayed in Fig. \ref{fig:junc_detail}~(c). Thus, the pair-wise geometric saliency of a junction $\jmath$ is defined as below, $\mathcal{N}$ is the amount of neighbors.
	\begin{equation}
	\omega_\jmath^{(2)} = \frac{1}{\mathcal{N}}\sum_{\jmath' \in \mathcal{N}_\jmath} e^{-\tau^{-1} \cdot \| \mathbf{c}_\jmath - \mathbf{c}_{\jmath'} \|_2} \cdot \omega_{\jmath'}^{(1)}.
	\label{eq:junction-index-add-neighbor}
	\end{equation}
	
	The geometric saliency of a VHR-RS image thus can be computed by summarizing $\omega_\jmath^{(1)}$ and $\omega_\jmath^{(2)} $ on each junctions, an example of which is shown in Fig. \ref{fig:junc_detail}~(d).

	\vspace{-3mm}
	\subsection{Geometric building index and building detection}
	\label{sec:gbi}
	
	Note that, given an $L$-junction $\jmath : \{\mathbf{c}, \vec \nu_1, \vec \nu_2, \rho \}$, the two branches $\vec \nu_1, \vec \nu_2$ uniquely form a parallelogram $R_\jmath$.
	Our {\it geometric building index} (GBI) attempts to associate each pixel $\mathbf{p}$ with a saliency measuring the possibility of the pixel belonging to buildings, which is the summation of saliency inside parallelogram of all junctions.
	Thus, for a pixel $\mathbf{p} \in \Omega$ in $U$, its corresponding GBI is calculated by:
	
	\begin{equation}
	\textrm{GBI}(\mathbf{p}) = \sum_{\jmath \in \mathcal{J}} \big( \omega_\jmath^{(1)} + \omega_\jmath^{(2)} \big) \cdot \,\mathbbm{1}_{\mathbf{p} \in R_\jmath},
	\label{eq:naive-gbi}
	\end{equation}
	where $J$ is the list of junctions detected by the ASJ detector in image $U$,
	and $\mathbbm{1}_{\mathbf{p} \in R_\jmath}$ is an indicator function, which equals $1$ if the pixel $\mathbf{p}$ is inside the parallelogram $R_\jmath$ of junction $\jmath$ and equals to $0$ otherwise.
	An illustration of the proposed GBI is shown in Fig. \ref{fig:junc_detail}~(e).
	
	For the image shown in Fig. \ref{fig:junc_detail}~(a), we simply threshold the computed GBI with its arithmetic average to finally generate the building map, as shown in Fig. \ref{fig:junc_detail}~(f).
	
	\vspace{-2mm}
	\section{Experiments and Discussions}
	\label{sec:experiment}
	This section evaluates the proposed method and compares it with state-of-the-art methods~\cite{shao_basi:_2014,huang_mbi_2012,liu2013perception,zuo_hf-fcn:_2016} on three public datasets that are used for validating building detection algorithms. The datasets are :
	\begin{itemize}
		\vspace{-2mm}
		\item[-] \emph{Spacenet-65 Dataset}\footnote{\url{https://amazonaws-china.com/cn/public-datasets/spacenet/}} consists of $65$ images of $2000 \times 2000$ pixels extracted from WorldView-2 satellite imagery, with a spatial resolution of $0.5$ meter. This dataset covers buildings in both urban and rural areas.
		\vspace{-2mm}
		\item[-] \emph{Potsdam Dataset}\footnote{\url{http://www2.isprs.org/commissions/comm3/wg4/2d-sem-label-potsdam.html}} contains $214$ images of $2000 \times 2000$ pixels with a spatial resolution of $0.05$ meter. Buildings in Potsdam exhibit to be large and are distributed dispersively due to the high resolution.
		\vspace{-2mm}
		\item[-] \emph{Massachusetts Buildings Dataset}\footnote{\url{https://www.cs.toronto.edu/~vmnih/data/}} contains $10$ images of $1500 \times 1500$ pixels in the test subset with a spatial resolution of $1$ meter.
		\vspace{-2mm}
	\end{itemize}
	
	To demonstrate the effectiveness of using geometric saliency in building detection, we compare our method with several state-of-the-art methods, including texture-based BASI~\cite{shao_basi:_2014}, morphology-based MBI~\cite{huang_mbi_2012}, local geometry-based PBI~\cite{liu2013perception} and learning-based HF-FCN~\cite{zuo_hf-fcn:_2016}.
	Note that BASI and PBI are designed for built-up area detection and the others aim to detect the accurate shape of buildings.
	For HF-FCN, we directly use the model provided by the authors.
	For quantitative evaluation, as in ~\cite{zuo_hf-fcn:_2016,automated2013}, the {\it mean Average Precision (mAP)} and {\it F-score} (also known as {\it F-measure}) are employed to measure the accuracy of detection.
	
	\vspace{-3mm}
	\subsection{Results and analysis}
	All results on the three datasets and detailed comparisons with different methods are available at \url{http://captain.whu.edu.cn/project/geosay.html}.
	Table \ref{table:map-f-datasets} shows the mAP and F-score of different building detection methods.
	It can be noted that the proposed GBI achieves the best performance in both mAP and F-score on {\it Spacenet-65} and {\it Potsdam} dataset, in the cases without training.
	When there are training samples, {\it i.e.,} the case on \emph{Massachusetts} dataset, HF-FCN outperforms all the other methods, since the model is fully trained on the dataset. But the model is severely overfitting, since it substantially loses its efficiency on {\it Spacenet-65} and {\it Potsdam} dataset and achieves very low mAP ($0.04$ and $0.03$) and F-score ($0.12$ and $0.10$).
	This questions the generalization capability of learning-based methods. By contrast, although the prior probabilities of junctions are estimated from {\it Spacenet-65} dataset, the high performance on both {\it Potsdam} and {\it Massachusetts} dataset indicates the powerful generality of our method.
	%In contrast, the worst performance of HF-FCN reflect its poor generality.
	%Indexes given by HF-FCN have a high accuracy with a low recall rate (see Fig.\ref{fig:visual_result}f), which suggests that the model trained from another dataset cannot fit well to images from a different domain.
	%Our method achieves robust performances and is robust to locate buildings with junctions.
	Even under the significant change of resolution (varying from 0.5m to 0.05m), the performance of our method is still better than the others.
	
	\begin{table}[htb!]
		\footnotesize
		\vspace{-3mm}
		\caption{Comparisons of different building detection methods in \textit{mAP} and \textit{F-score}. Note that our method outperforms the others in the cases without training.}
		\vspace{-4mm}
		\begin{center}
			\begin{tabular}{c|c|c|c|c|c|c}
				\hline
				\multirow{2}{*}{Method}&
				\multicolumn{2}{c|}{Spacenet-65}&\multicolumn{2}{c|}{Massachusetts}&\multicolumn{2}{c}{Potsdam}
				\cr\cline{2-7}&mAP&F-score &mAP&F-score &mAP&F-score \cr
				\hline
				BASI~\cite{shao_basi:_2014} & 0.34 & 0.44 & 0.32 & 0.40 & 0.34 & 0.44\\
				MBI~\cite{huang_mbi_2012} & 0.28 & 0.35 & 0.28 & 0.38 &  0.17 & 0.35\\
				%Pantex & 0.2074 & 0.3088 & 0.2904 & 0.3758  & 0.2704 & 0.3974\\
				PBI~\cite{liu2013perception} & 0.27 & 0.37 & 0.25 & 0.36  & 0.41 & 0.50\\
				HF-FCN~\cite{zuo_hf-fcn:_2016} & 0.04 & 0.12 & \textbf{0.57} & \textbf{0.74} & 0.03 & 0.10\\
				%			RC & 0.1970 & 0.2765  & 0.2639 & 0.3479 & 0.4081 & 0.5098\\
				GBI (ours) & \textbf{0.46} & \textbf{0.52} & 0.37 & 0.44 & \textbf{0.46} & \textbf{0.59}\\
				\hline
			\end{tabular}
		\end{center}
		\label{table:map-f-datasets}
		\vspace{-4mm}
	\end{table}

	\begin{figure*}[htb!]
		\vspace{-3mm}
		\centering
		\includegraphics[width = 0.92\linewidth]{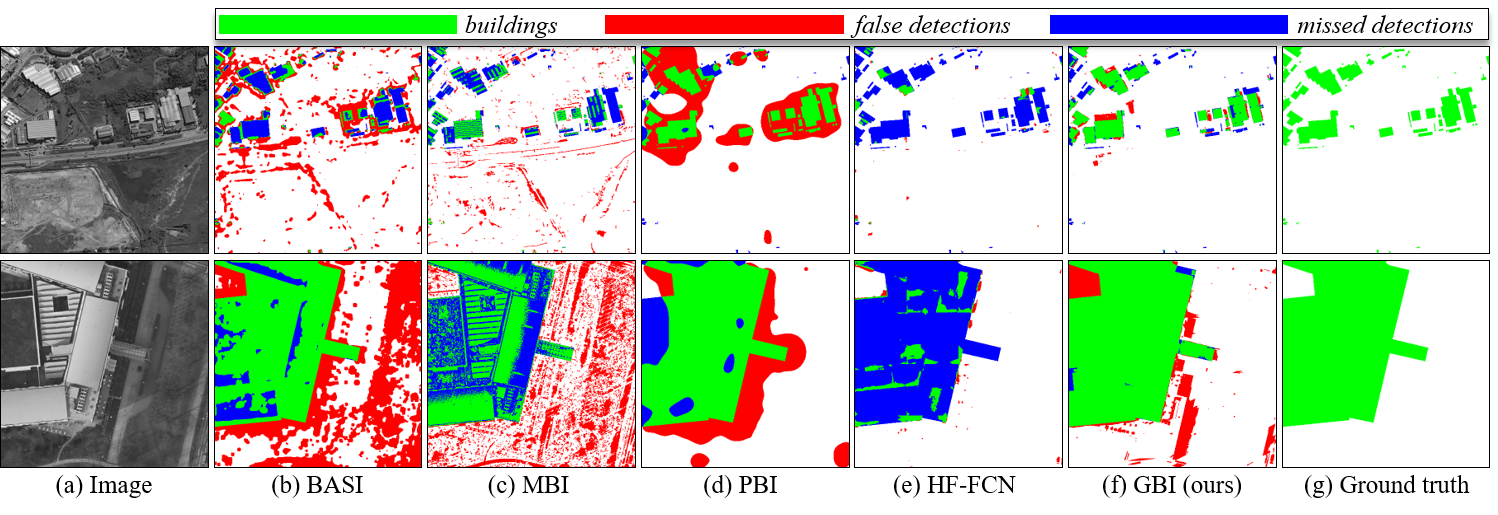}
		\vspace{-4mm}
		\caption{Building detection results on three sample images. On the first image that has many roads and empty areas, our method highlights buildings, while the compared methods detect a lot of redundant non-building areas.
			On the second image, BASI performs poorly due to the lack of texture and MBI generates many failures because of the low contrast between buildings and background. More results in \url{http://captain.whu.edu.cn/project/geosay.html}.
		}
		\label{fig:visual_result}
		\vspace{-3mm}
	\end{figure*}
	
	Fig.~\ref{fig:visual_result} illustrates the building detection results on two sample images.
	The first image shows a case where buildings are distributed dispersively and a lot of non-building objects exist.
	The two built-up area detection methods, PBI and BASI, extract not only the buildings but also the neighbors, and produce many failures.
	The texture-based BASI results in numerous false detections in textural regions like roads and forest, and the local geometry-based PBI results in a lot of false detections around buildings.
	Other methods like MBI also face such problem, which confuse the rural roads with buildings.
	The phenomenons above suggest that the building indexes defined by these methods are not suitable to describe buildings in VHR-RS images.
	For the second image, BASI misses many parts of the three highlight buildings with low-texture roofs, which indicates that texture-based methods are inappropriate to buildings with low textures.
	MBI detects most of the buildings but fails to extract the whole shape of the central building due to the imbalanced luminance at the roof.
	By contrast, such cases do not hamper the performance of our method, since junctions locate at the corners of buildings no matter what the texture or the luminance of buildings appear.
	\vspace{-4mm}
	\subsection{Discussions}
	The proposed GBI is based on the geometric saliency in VHR-RS images, not requiring any annotated training samples for the computations, and is capable of preserving the whole geometric shapes of buildings with high performance. Such results are promising for mapping buildings in VHR-RS images.
	One limitation of the GBI is that, in VHR-RS images, some man-made architectures or objects (e.g., cars) may also exhibit salient geometrical structures, which may lead to false detections. For solving these problems, some prior information in the images,  such as ratios between object size and image resolution, can be used to suppress false alarms.
	In addition, it is also of great interest to incorporate different kinds of information to improve the detection accuracy of the position and whole boundaries of buildings.
	
	\vspace{-3mm}
	\section{Conclusion}
	\label{sec:conclusion}
	\vspace{-2mm}
	This paper proposes a geometric saliency-based method for detecting buildings in VHR-RS images. Compared with traditional saliency-based methods, our method measures the geometric saliency of building by leveraging the meaningful geometric features that are specialized for describing buildings; compared with the learning-based method, our method is totally unsupervised and free of any training strategies. Experiments on three public datasets demonstrate that the proposed method not only achieves a substantial performance improvement, but also generalizes well to data of broad domains.
	Moreover, the buildings detected by our method have a clearer boundary and less redundant cluttered areas than existing methods.
	
	%\vfill
	%\pagebreak
	\stepcounter{section}
	\renewcommand\refname{\centering \normalsize \thesection. REFERENCES \vspace{-2mm}}
	\footnotesize
	\bibliographystyle{IEEEbib}
	\bibliography{bib/refs}

\begin{thebibliography}{10}

\bibitem{zha_use_2003-1}
Y.~Zha, J.~Gao, and S.~Ni,
\newblock ``Use of normalized difference built-up index in automatically
  mapping urban areas from {{TM}} imagery,''
\newblock {\em International Journal of Remote Sensing}, vol. 24, no. 3, pp.
  583--594, 2003.

\bibitem{pesaresi_robust_2008}
M.~Pesaresi, A.~Gerhardinger, and F.~Kayitakire,
\newblock ``A robust built-up area presence index by anisotropic
  rotation-invariant textural measure,''
\newblock {\em IEEE Journal of Selected Topics in Applied Earth Observations
  and Remote Sensing}, vol. 1, no. 3, pp. 180--192, 2008.

\bibitem{shao_basi:_2014}
Z.~Shao, Y.~Tian, and X.~Shen,
\newblock ``{{BASI}}: A new index to extract built-up areas from
  high-resolution remote sensing images by visual attention model,''
\newblock {\em Remote Sensing Letters}, vol. 5, no. 4, pp. 305--314, 2014.

\bibitem{sirmacek_urban_2009}
B.~Sirmacek and C.~Unsalan,
\newblock ``Urban area detection using gabor features and spatial voting,''
\newblock in {\em 2009 {{IEEE}} 17th {{Signal Processing}} and {{Communications
  Applications Conference}}}, 2009, pp. 812--815.

\bibitem{sirmacek_urban_2010}
B.~Sirmacek and C.~Unsalan,
\newblock ``Urban area detection using local feature points and spatial
  voting,''
\newblock {\em IEEE Geoscience and Remote Sensing Letters}, vol. 7, no. 1, pp.
  146--150, 2010.

\bibitem{huang_mbi_2012}
X.~Huang and L.~Zhang,
\newblock ``Morphological building/shadow index for building extraction from
  high-resolution imagery over urban areas,''
\newblock {\em IEEE Journal of Selected Topics in Applied Earth Observations
  and Remote Sensing}, vol. 5, no. 1, pp. 161--172, 2012.

\bibitem{saito_multiple_2016}
S.~Saito, T.~Yamashita, and Y.~Aoki,
\newblock ``Multiple object extraction from aerial imagery with convolutional
  neural networks,''
\newblock {\em Electronic Imaging}, vol. 2016, no. 10, pp. 1--9, 2016.

\bibitem{zuo_hf-fcn:_2016}
T.~Zuo, J.~Feng, and X.~Chen,
\newblock ``{HF}-{FCN}: Hierarchically fused fully convolutional network for
  robust building extraction,''
\newblock in {\em Asian Conference on Computer Vision}, 2016, pp. 291--302.

\bibitem{xia-stucture-2010}
G.-S. Xia, W.~Yang, J.~Delon, Y.~Gousseau, H.~Sun, and H.~Ma{\^i}tre,
\newblock ``{Structural High-resolution Satellite Image Indexing},''
\newblock in {\em {ISPRS TC VII Symposium - 100 Years ISPRS}}, Vienna, Austria,
  2010, vol. XXXVIII, pp. 298--303.

\bibitem{Xue2017Anisotropic}
N.~Xue, G.~Xia, X.~Bai, L.~Zhang, and W.~Shen,
\newblock ``Anisotropic-scale junction detection and matching for indoor
  images,''
\newblock {\em IEEE Trans. Image Processing}, vol. PP, no. 99, pp. 1--1, 2017.

\bibitem{liu2013perception}
G.~Liu, G.~Xia, X.~Huang, W.~Yang, and L.~Zhang,
\newblock ``A perception-inspired building index for automatic built-up area
  detection in high-resolution satellite images,''
\newblock in {\em IGARSS 2013}, pp. 3132--3135.

\bibitem{automated2013}
A.~O. Ok, C.~Senaras, and B.~Yuksel,
\newblock ``Automated detection of arbitrarily shaped buildings in complex
  environments from monocular vhr optical satellite imagery,''
\newblock {\em IEEE Transactions on Geoscience and Remote Sensing}, vol. 51,
  no. 3, pp. 1701--1717, 2013.

\end{thebibliography}
	
\end{document}